\documentclass[letterpaper, 10 pt, conference]{ieeeconf}  

\IEEEoverridecommandlockouts                              
\usepackage[pdftex]{graphicx}
\usepackage[cmex10]{amsmath}
\usepackage{amssymb}
\usepackage{import}
\usepackage{bm}

\usepackage{color}
\usepackage{units}
\usepackage{url}
\usepackage{float}
\usepackage{epstopdf}
\usepackage{stmaryrd}

\usepackage{units}
\usepackage{siunitx}

\usepackage{outlines}

\usepackage[utf8]{inputenc}
\usepackage{boldline}
\usepackage[flushleft]{threeparttable}

\usepackage[doi=false,isbn=false,defernumbers=true,style=ieee,backend=bibtex,maxbibnames=1000]{biblatex}

\usepackage{mathtools}
\usepackage{multirow}

\newcommand{\mnorm}[1]{\left\| #1 \right\|}

\newcommand{\mvec}[1]{\boldsymbol{#1}}

\newcommand{\matSym}[1]{\boldsymbol{#1}}

\newcommand{\mmat}[1]{\begin{bmatrix} #1 \end{bmatrix} }

\newcommand{\mrb}[1]{\left( #1 \right)} 

\newcommand{\mnoshow}[1]{}

\newcommand{\figref}[1]{Fig.~\ref{#1}}
\newcommand{\secref}[1]{Section~\ref{#1}}

\newcommand{\ddt}{\frac{\mathrm{d}}{\mathrm{d}t}}

\newcommand*{\transpose}{T}

\title{\LARGE \bf
An algorithm for real-time restructuring of a ranging-based
localization network
}

\author{Saman Fahandezh-Saadi and Mark W. Mueller
\thanks{The authors are with the Department of Mechanical Engineering, at the University of California, Berkeley.
\newline        {\tt \{samanfahandej,mwm\}@berkeley.edu}}%
}

\begin{document}

\maketitle
\thispagestyle{empty}
\pagestyle{empty}

\begin{abstract}
This paper presents a method to improve the localization accuracy of robots operating in a range-based localization network. The method is  favorable especially when the robots operate in harsh environments where the access to a robust and reliable localization system is limited. A state estimator is used for a six degree of freedom object using inertial sensors as well as an Ultra-wideband (UWB) range measurement sensor. The estimator is incorporated into an adaptive algorithm, improving the localization quality of an agent by using a mobile UWB ranging sensor, where the mobile anchor moves to improve localization quality. The algorithm reconstructs localization network in real-time to minimize the determinant of the covariance matrix in the sense of least square error. Finally, the proposed algorithm is experimentally validated in a network consisting of one mobile and four fixed anchors.
\end{abstract}

\section{Introduction}

An accurate, robust, and accessible localization system is crucial for robot operation in the case of, for example, emergency services, building fault detection in disaster areas, or rescue missions. We assume that in these situations the pre-installed infrastructure is destroyed or limited (i.e. no access to GPS signal), and therefore a local and stand-alone alternative localization system is vital. Also, the harsh environment of these situation requires to use reliable sensory devices. For example, the sensors should function in smoke, dust, and in foggy or heavy rain conditions where optical systems cannot be used reliably. An infrastructure (e.g. radio/audio beacons) for the local positioning system is important to be placed in the environment such that it maximizes the localization accuracy, but it is most likely impossible due to limited accessibility of such environment. The proposed method in this paper will address above issues by using a reliable range measurement sensor with mobile infrastructure setting.

The recently popular ultra-wideband radio ranging is a flexible, relatively low-cost, and reliable localization technology mostly for but not limited to indoor environments -- see e.g. \cite{UWBdustSmoke, UWBRescueMission, SahinogluUWBBook, uwbTechnology, Mahfouz2008Investigation}. The infrastructure components (i.e radio-beacon) of the system are easy to setup in the environment. The radio-beacon (here called `anchor') communicates via radio messages with an agent which is equipped with UWB sensor. The result is a range measurement, the distance between the agent and each individual anchors. 

The UWB sensors are used in a wide range of localization methods. For example, in conjunction with the SLAM (simultaneous localization and mapping) problem, a particular UWB transmitter-receiver configuration on an agent is used in \cite{UWB_SLAM2010}. The authors explain that it is crucial to use UWB ranging specially in emergency situation, when other technologies like camera- or laser-based sensors fail to operate in such harsh environment. The UWB technology also has medical applications. For example, authors in \cite{surgicalapp2017} developed an algorithm to improve the localization accuracy of surgical devices using UWB sensor in highly reflective and dense indoor environments such as operating rooms where multipath and no-line-of-sight conditions are an issue. In \cite{surgicalapp2010}, authors developed a Kalman filter estimator using fusion of inertial measurement unit with UWB ranging sensor for localization in crowded and dynamic environments like imaging rooms, where technical devices like C-arm imaging systems or operating tables are moving.   

\begin{figure}
    \centering
    \includegraphics{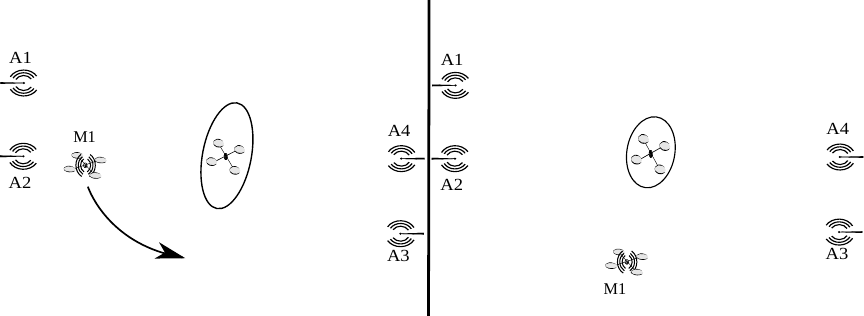}
    \caption{
    A schematic of the proposed systems: an agent (here a quadcopter) operates in a space prepared with combination of one mobile (indexed as M1) and four fixed radio anchors (indexed A1-A4). At the left, the ellipse representing the position uncertainty is extended in the direction which there is no anchors present. At the right, the mobile anchor moves to the area with no anchors and reduces the uncertainty around the agent. 
    }
    \label{fig:sysLayout}
\end{figure}

The range measurement based localization is also used in multi-agent cooperative positioning and wireless sensor network \cite{firefighter_localization2014, coop_localization2002, localization_wifi2006, Wang2015, multiagentLocCamera}. The main idea is to localize several agents (or wireless sensor nodes) with the capability of exchanging their information (i.e. relative distance, position, orientation, etc). Our proposed method considers using several fixed UWB anchors as well as mobile anchors, the ones that adapt their positions in order to improve the position accuracy of the agent. In other words, the cooperation occurs between an agent and a set of mobile anchors. Since the localization accuracy is sensitive to the anchor arrangement, it is hard to achieve reasonable accuracy with restricted anchor's arrangement without careful planning, which may be difficult in harsh conditions. In this paper, the position of mobile anchors is assumed to be known (i.e. they have access to a separate, independent localization technology), whereas the agents rely only on the mobile and fixed anchors for localization, since they don't have access to any localization infrastructure. Future work will look at generalizing this by removing the separate localization technology which mobile anchors currently rely on. A schematic of our proposed system is shown in \figref{fig:sysLayout}

In this paper, we use a state estimator with no assumption on force-torque, and we develop an adaptive localization algorithm on top of the estimator for moving a known ranging measurement point (i.e. a mobile anchor), so that an agent may minimizes its position uncertainty. The state estimation consists of a general six degrees of freedom (6DOF) model with measurement model as agent's distance to known positions in the world. The adaptive algorithm maximizes the agent's localization quality in the least-square sense. The results are validated in a series of experiments, where the estimator is implemented on a quadcopter system.  

\section{System Model}
\label{secModel}
We consider model of a generic 6 degree of freedom (3 in translation and 3 in rotation) rigid body for the purpose of estimating the states of an agent. The agent is equipped with inertial measurement unit (accelerometer and rate gyroscope) and a range measurement sensor that allows to measure the distance to any of fixed or mobile sets of anchors (with known location) in its environment. In the following subsections we will briefly describe the underlying models of the estimator. 
\subsection{Equations of motion}
We will use the convention of using bold-face symbols for vector/matrix quantities, and regular font for scalars. In the model, the position of the rigid body (agent) is denoted as $\mvec{x}$, its velocity as $\mvec{v}$, and acceleration as $\mvec{a}$, all expressed in the inertial frame fixed to the ground. The rotation matrix and the angular velocity are given respectively by $\mvec{R}$ and $\mvec{\omega}$, where the rotation matrix represents the orientation of the agent. The multiplication by the roation matrix will result a coordination transformation from the body-fixed frame to the inertial frame. The time derivatives of these quantities are given as 
\begin{align}
   \ddt \mvec{x} &= \mvec{v} 
\\ \ddt \mvec{v} &= \mvec{a}
\\ \ddt \mvec{R} &= \mvec{R}\mvec{S(}\mvec{\omega)}
\end{align}
note that $\mvec{S(}\mvec{\omega)}$ is the skew-symmetric matrix version of cross product, where $\mvec{S(}\mvec{x)}\mvec{y} = \mvec{x}\times\mvec{y}$. 
\subsection{Inertial measurements}
The inertial measurement unit outputs the accelerometer and rate gyroscope measurements, $\mvec{\alpha}$ and $\gamma$. The accelerometer measures the `proper acceleration' in the body-fixed frame which we assume it is corrupted by additive noise $\mvec{\nu}_{\alpha}$.
\begin{align}
	\mvec{\alpha}_m = \mvec{R}^{-1} \mrb{\mvec{a} - \mvec{g}} + \mvec{\nu}_\alpha
\end{align}
The gravitational acceleration is in the inertial frame fixed to the ground, and has magnitude $\mvec{g}=\mrb{0,0,-9.81}\si{m\per s\squared}$. The rate gyroscope measures angular velocity of the agent in the body-fixed frame. The measurement is modeled as    
\begin{align}
\mvec{\gamma} = \mvec{\omega} + \mvec{\nu}_\gamma
\end{align}
here the the measurement is corrupted by $\mvec{\nu}_{\gamma}$. Both $\mvec{\nu}_{\alpha}$ and $\mvec{\nu}_{\gamma}$ are assumed to be zero mean, based on the fact that the sensors are well calibrated and scale/bias-free.
\subsection{UWB Range measurement system}
The UWB radio mounted on the agent communicates with other radios in the environment. At each time instant, the agent measures the distance from its position at $\mvec{x}$ to one of fixed or mobile position at $\mvec{p}_i$ (anchor's position) in the world. This measurement $\rho_i$ to anchor $i$ is modelled as the Euclidean norm corrupted with additive scalar noise $\nu_{\rho}$ with zero mean.
\begin{align}
    \rho_i = \mnorm{\mvec{x} - \mvec{p}_i} + \nu_\rho \label{eqModelRangeMeasurent}
\end{align}
The UWB radio uses two-way ranging time-of-flight based algorithm to calculate the distance (see \cite{mueller2015fusing} for two-way ranging scheme). The agent can communicate only with one anchor at a time, meaning that only a single range measurement can be taken at any instant in time.

\section{State Estimator}
\label{secEstimator}

For the state estimation, an extended Kalman filter (EKF) \cite{simon2006optimal} presented in this section estimates the 12-element state of the agent consist of position, attitude and their derivatives. The kinematic model makes the Kalman filter an estimator for a generic 6DOF rigid body with no assumption on forces or torques acting on the agent. The EKF uses the technique of \cite{mueller2016covariance} to encode an attitude in the state with correct-to-first-order statistics. 


The estimator does not include the angular velocity as a state, instead uses the output of rate gyroscope measurement; assuming that the modern sensors output high-quality measurements which is a standard approach in attitude estimation for satellites \cite{markley2003attitude}. Thus, the estimator's stochastic state $\mvec{\xi}$ is a 9 dimensional vector:
\begin{align}
    \hat{\mvec{\xi}} = \mrb{\hat{\mvec{x}}, \hat{\mvec{v}}, \hat{\mvec{\delta}}}
\end{align}
with the hat denoting estimated quantities, and where $\hat{\mvec{\delta}}$ represents attitude error measure, assumed to be small. The estimator uses a redundant attitude representation, with a `reference attitude' $\mvec{R}_\mathrm{ref}$ and the attitude error $\hat{\mvec{\delta}}$ combined yielding the estimator's attitude estimate $\hat{\mvec{R}}$
\begin{align}
    \hat{\mvec{R}} = \mvec{R}_\mathrm{ref} \mrb{\mvec{I} + \mvec{S(}\hat{\mvec{\delta}}\mvec{)}}
\end{align}
with $\mvec{I}$ the identity matrix. This representation allows for a singularity-free attitude estimation using only a three-dimensional representation of the attitude error -- a complete discussion of this approach is given in \cite{mueller2016covariance}. 

The EKF uses the output of accelerometer and rate gyroscope for the prediction step, so the state differential is given by
\begin{align}
    \ddt \mvec{v} &= \mvec{R} \mvec{\alpha}_m + \mvec{g} + \mvec{R}\mvec{\nu}_\alpha \\
    \ddt \mvec{\delta} &= \mvec{\gamma} - \mvec{\nu}_\gamma
\end{align}
In the measurement update step, the estimator uses the output of the UWB ranging radios. The linearization of measurement equation \eqref{eqModelRangeMeasurent} is as follows
\begin{align}
    \mvec{H}_i :=& \frac{\partial \rho_i}{\partial \mvec{\xi}}  = \mrb{\frac{\partial \rho_i}{\partial \mvec{x}}, \frac{\partial \rho_i}{\partial \mvec{v}}, \frac{\partial \rho_i}{\partial \mvec{\delta}}} \label{eqMeasPartialEqn}
\\  \frac{\partial \rho_i}{\partial \mvec{x}} =& \frac{\mvec{x}-\mvec{p}_i}{\mnorm{\mvec{x}-\mvec{p}_i}} =: \mvec{e}_i  \label{eqUnitVec}
\\  \frac{\partial \rho_i}{\partial \mvec{v}} =& \frac{\partial \rho_i}{\partial \mvec{\delta}} = \mvec{0}
\end{align}
Note that the measurement sensitivity with respect to the agent's position $\mvec{e}_i$ is the unit vector in the direction of anchor $i$ from the agent.

The estimate covariance matrix $\mvec{\Sigma}$ computed by EKF relies on the partial derivatives of linearization process and using the approach of \cite{mueller2016covariance}. This is the partitioned matrix as below 
\begin{align}
\matSym{\Sigma} = 
	\begin{bmatrix}	
		\matSym{\Sigma_{\mvec{x}\mvec{x}}} & \matSym{\Sigma_{\mvec{x}\mvec{v}}} & \matSym{\Sigma_{\mvec{x}\mvec{\delta}}}\\
		\matSym{\Sigma_{\mvec{x}\mvec{v}}^T} & \matSym{\Sigma_{\mvec{v}\mvec{v}}} & \matSym{\Sigma_{\mvec{v}\mvec{\delta}}}\\
		\matSym{\Sigma_{\mvec{x}\mvec{\delta}}^T} & \matSym{\Sigma_{\mvec{v}\mvec{\delta}}^T} & \matSym{\Sigma_{\mvec{\delta}\mvec{\delta}}}\\
	\end{bmatrix} \in \mathbb{R}^{9\times9}
\end{align} 
with e.g. $\mvec{\Sigma}_{\mvec{x}\mvec{\delta}}$ the $3\times3$ cross-covariance between the position and attitude states. The very intuitive property of measurement sensitivity will be used in the following section to determine how the anchors can move in order to maximize the localization quality of the agent.

\section{Mobile anchor algorithm} \label{secMobileAlgorithm}

The measurement model as described in \eqref{eqUnitVec} is sensitive to the location of the anchors $\mvec{p}_i$. This means, it is possible to affect the variance of the state estimates by moving the anchors. This allows to create an optimization problem in order to move the anchors in the direction which minimizes the estimation error of the agent's position. The covariance matrix is used as the metric for estimation quality. In the following subsections we will discuss this approach in detail.   
\subsection{Least squares approach}
The least-square approach is used as an easy-to-analyze approximation of the EKF used to estimate the agent's location. This is motivated by the least squares interpretation of the Kalman filter.

All the derivations in this section are assumed to be for a single mobile anchor with $N-1$ fixed anchors. It is simple to generalize the derivation for multiple mobile anchors, since the desired moving direction of each anchor decouples. All the quantities without number subscription are introduced for the single mobile anchor. 

Notable is that we have two sets of decision variables in this section. The position estimate of the agent $\mvec{x}$ is the decision variable for the least squares problem. The mobile anchor's position $\mvec{p}$ is the decision variable for minimizing the variance of the agent's position estimate. We are using these two variables throughout this section. 

We consider again the ranging measurement model in \eqref{eqModelRangeMeasurent}. Note, here we look at a batched version of this equation (i.e. the equation is concatenation of scalar measurements at each discrete time step for $N$ anchors, fixed or mobile anchors), which is an approximation of EKF
\begin{align}
    \mvec{\rho} &= \mvec{h}(\mvec{x}, \mvec{p}) + \mvec{\nu}_{\rho} \label{eqModelRangeMeasurentBatch}\\
    \mmat{\rho\\\rho_1\\ \vdots \\ \rho_{N-1}} &= \mmat{\mnorm{\mvec{x} - \mvec{p}}\\ \mnorm{\mvec{x} - \mvec{p}_1} \\ \vdots \\ \mnorm{\mvec{x} - \mvec{p}_{N-1}}} + \mmat{\nu_{\rho}\\\nu_{\rho_1}\\ \vdots \\ \nu_{\rho_{N-1}}}, \nonumber  
\end{align}
where $\mvec{\rho}$ is the measurement vector, $\mvec{\nu}_{\rho}$ is the additive noise vector with zero mean, and $\mvec{h}(\mvec{x, p})$ is the nonlinear measurement model (vector-valued equation) mapping the position of an agent in $\mathbb{R}^3$ to the batch of anchors' measurements in $\mathbb{R}^N$. The partial derivative of the nonlinear model with respect to the agent's position will result a $N \times 3$ Jacobian matrix
\begin{align}
    \mvec{A}(\mvec{\hat{x}, p}) = \frac{\partial \mvec{h}(\mvec{x, p})}{\partial \mvec{x}} = \mmat{\mvec{e}^{\transpose} \\ \mvec{e}_1^{\transpose} \\ \vdots \\ \mvec{e}_{N-1}^{\transpose}} \label{eq:linearizedMeasModel} \in \mathbb{R}^{N\times 3}
\end{align}
where $i$th row of the matrix is the unit vector pointing from the agent to anchor $i$. Assuming noise in \eqref{eqModelRangeMeasurentBatch} to be zero mean with isotropic covariance $\mathrm{Var}(\mvec{\nu}_{\rho})=q\mvec{I}$, the position estimate can be found by minimizing the $2-$norm squared of noise
\begin{align*}
    &\mvec{\hat{x}} = \textnormal{arg}\ \min_{\mvec{x}} \mnorm{\mvec{\nu}_{\rho}}^{2}
\end{align*}
As stated before, for the nonlinear measurement model this will be done iteratively by linearizing the model at each time step, and moving in the direction of the gradient descent. 
\subsection{Statistical properties}
Since we use the linearized version of the equation, the mean of the estimated position is zero (i.e. unbiased estimator) but the variance is
\begin{align}
    \mathrm{Var}(\mvec{\hat{x}}) &\approx ((\mvec{A}^{\transpose} \mvec{A})^{-1} \mvec{A}^{\transpose}) q\mvec{I} ((\mvec{A}^{\transpose} \mvec{A})^{-1} \mvec{A}^{\transpose})^{\transpose}  \nonumber\\
    &= q(\mvec{A}^{\transpose} \mvec{A})^{-1}
\end{align}
For brevity, we use $\mvec{A}$ instead of $\mvec{A}(\mvec{\hat{x}, p})$ from now on. Our goal is to minimize the effect of noise on the state estimates (i.e. to minimize the variance of the state estimates by varying the anchor location). Specifically, we choose to minimize the largest eigenvalue of the covariance matrix, which is equivalent to maximizing the smallest eigenvalue of its inverse:
\begin{align}
    \max_{\mvec{p}} \ \min_{i} \ \lambda_i((\mvec{A}^{\transpose} \mvec{A})) \label{eq:maxMin}
\end{align}
By expanding matrix $\mvec{A}^{\transpose} \mvec{A}$, the max-min problem can be converted to a simple form. Substituting \eqref{eq:linearizedMeasModel} in matrix $\mvec{A}$, the result is the sum of $N$ rank-one matrices
\begin{align}
    \mvec{A}^{\transpose} \mvec{A} &=  \mmat{\mvec{e} & \mvec{e}_1 & \hdots & \mvec{e}_{N-1}} \mmat{\mvec{e}^{\transpose} \\ \mvec{e}_1^{\transpose} \\ \vdots \\ \mvec{e}_{N-1}^{\transpose}} \\ \nonumber
    &= \mvec{e} \mvec{e}^{\transpose} + \sum_{i=1}^{N-1} \mvec{e}_i \mvec{e}_i^{\transpose} \label{eq:rankOne}
\end{align}
Since all $\mvec{e}_i$ and $\mvec{e}$ are unit vectors (the direction to the anchors), the trace of $\mvec{A}^{\transpose}\mvec{A}$ is always constant and equal to the number of anchors:  
\begin{align}
    \mathrm{tr}(\mvec{A}^{\transpose} \mvec{A}) &= \mathrm{tr}(\mvec{e}^{\transpose} \mvec{e}) + \sum_{i=1}^{N-1} \mathrm{tr}(\mvec{e}_i^{\transpose} \mvec{e}_i) = N
\end{align}
The trace of a matrix is also equal to the sum of its eigenvalues; that means the sum of eigenvalues of the covariance matrix in this problem is constant.
\begin{align}
    \mathrm{tr}(\mvec{A}^{\transpose} \mvec{A}) = \sum_{i=1}^{3} \lambda_i = N 
    \label{eq:sumofeig}
\end{align}
Using this fact, the max-min optimization problem can be transformed to 
\begin{equation} 
    \begin{aligned}
        & \underset{\mvec{\lambda}, t}{\text{max}} \ \ t \\
        & \text{subject to} &  \mvec{1}^{\transpose}\mvec{\lambda} = N \\
        & & t \leq \lambda_i \ \ \forall i \\
        & & 0 \leq \lambda_i \ \ \forall i 
    \end{aligned}
    \label{eq:slack}
\end{equation}
where $t$ is a slack variable, and $\mvec{\lambda}$ is a vector representing the eigenvalues of matrix $\mvec{A}^{\transpose} \mvec{A}$. This problem can be solved by introducing its dual problem. Since \eqref{eq:sumofeig} shows that the sum of eigenvalues is constant, we can conclude that \eqref{eq:slack} is equivalent to the following problem
\begin{align}
    \max_{\mvec{p}} \ \mathrm{det}(\mvec{A}^{\transpose} \mvec{A}) =  
    \max_i \ \prod_{i=1}^{3} \lambda_i \label{eq:maxDet}
\end{align}
It can be shown that for both problems the maximum occurs, when all the eigenvalues are equal. Specifically, \eqref{eq:maxMin} has interpretation for robust design (it minimizes the worst variance direction) as opposed to \eqref{eq:maxDet} which minimizes the volume of ellipsoid associated with the covariance matrix \cite{Boyd:2004:CO:993483}. This shows that this method satisfies both design approaches (i.e. minimizing the largest variance direction will result in minimizing the ellipsoid volume). 

\subsection{Mobile anchor}
Since the matrix consists of the mobile anchor's position, its determinant depends on the position of the mobile anchor at each time step.  
\begin{align*}
\matSym{A^{\transpose}A} &= \frac{(\mvec{\hat{x}} - \mvec{p})(\mvec{\hat{x}} - \mvec{p})^{\transpose}}{\mnorm{\mvec{\hat{x}} - \mvec{p}}} \\
&+ \sum_{i=1}^{N-1} \frac{(\mvec{\hat{x}} - \mvec{p}_i)(\mvec{\hat{x}} - \mvec{p}_i)^{\transpose}}{\mnorm{\mvec{\hat{x}} - \mvec{p}_i}} \in \mathbb{R}^{3\times 3}
\end{align*}
By taking partial derivative with respect to the mobile anchor's position $\mvec{p} = (x, y, z)$ we can find a direction that the mobile anchor can move in order to minimize the determinant of the covariance matrix
\begin{align}
    \frac{\partial \mathrm{det}(\mvec{A}^{\transpose} \mvec{A})}{\partial \mvec{p}} &= 
    \mmat{\frac{\mathrm{d}}{\mathrm{d} x} \mathrm{det}(\mvec{A}^{\transpose} \mvec{A})\\
    \frac{\mathrm{d}}{\mathrm{d} y} \mathrm{det}(\mvec{A}^{\transpose} \mvec{A}) \\ 
    \frac{\mathrm{d}}{\mathrm{d} z} \mathrm{det}(\mvec{A}^{\transpose} \mvec{A})} \label{eq:derivativeOfdet}
\end{align}
We rewrite the inverse of the covariance matrix in a compact form as
\begin{align}
    \mvec{A}^{\transpose}\mvec{A} = \mmat{\mvec{m}_1 & \mvec{m}_2 & \mvec{m}_3\\
                                          \mvec{m}_2 & \mvec{m}_4 & \mvec{m}_5\\
                                          \mvec{m}_3 & \mvec{m}_5 & \mvec{m}_6} \label{eq:mform}
\end{align}
where $\mvec{m}_i$ are functions of mobile anchor's position. As an example, $\mvec{m}_1$ and $\mvec{m}_6$ have the form (note that $\hat{x}_1$, $\hat{x}_2$, $\hat{x}_3$ are the three first states of the estimator representing the agent's position):
\begin{align*}
    \mvec{m}_2 = \frac{(\hat{x}_1 - x)(\hat{x}_2 - y)}{\mnorm{\mvec{\hat{x}}-\mvec{p}}^2}, \quad
    \mvec{m}_6 = \frac{(\hat{x}_3 - z)^2}{\mnorm{\mvec{\hat{x}}-\mvec{p}}^2}
\end{align*}
Representing the inverse of the covariance matrix in the compact form of \eqref{eq:mform} allows to derive the determinant of the matrix and its partial derivative in terms of functions $\mvec{m}_i$, and then substitute back the mobile anchor's position. At the end, \eqref{eq:derivativeOfdet} will be a set of three closed form equations in terms of the mobile anchor's position as well as the agent's position, which can be easily implemented.

We choose the control action as the velocity command for the mobile anchor which is the Jacobian of the determinant of the covariance matrix multiplying by a gain value $s > 0$ as follows
\begin{align}
    \mvec{v}_{cmd} &= s \frac{\partial \mathrm{det}(\mvec{A}^{\transpose} \mvec{A})}{\partial \mvec{p}} \nonumber
\end{align}

 This is exactly the same as moving on the direction of gradient ascent iteratively until the mobile anchor reaches to the top (max point of determinant) and stays there. For sequence of steps the mobile anchors move as follows  
\begin{align}\label{eq:strategy}
    \mvec{p}(k+1) = \mvec{p}(k) + (\Delta t) \mvec{v}_{cmd}
\end{align}
where $k$ is time steps and $\Delta t$ is the discrete time interval. In the following section we will present the real-time experimental results of this approach. 

\section{Experimental validation}
\label{secExpValidation}
The approach is validated in experiment, where two quadcopters are used, one as autonomous agent and one as mobile anchor. A first set of experiments just uses the fixed anchors as the only UWB measurement setting. The second experiment uses one mobile anchor in addition to the fixed anchors. Results will be presented by comparing the agent's variance computed by the onboard EKF for the two different experiment settings.
\begin{figure*}
    \centering
    \makebox[0pt]{\includegraphics{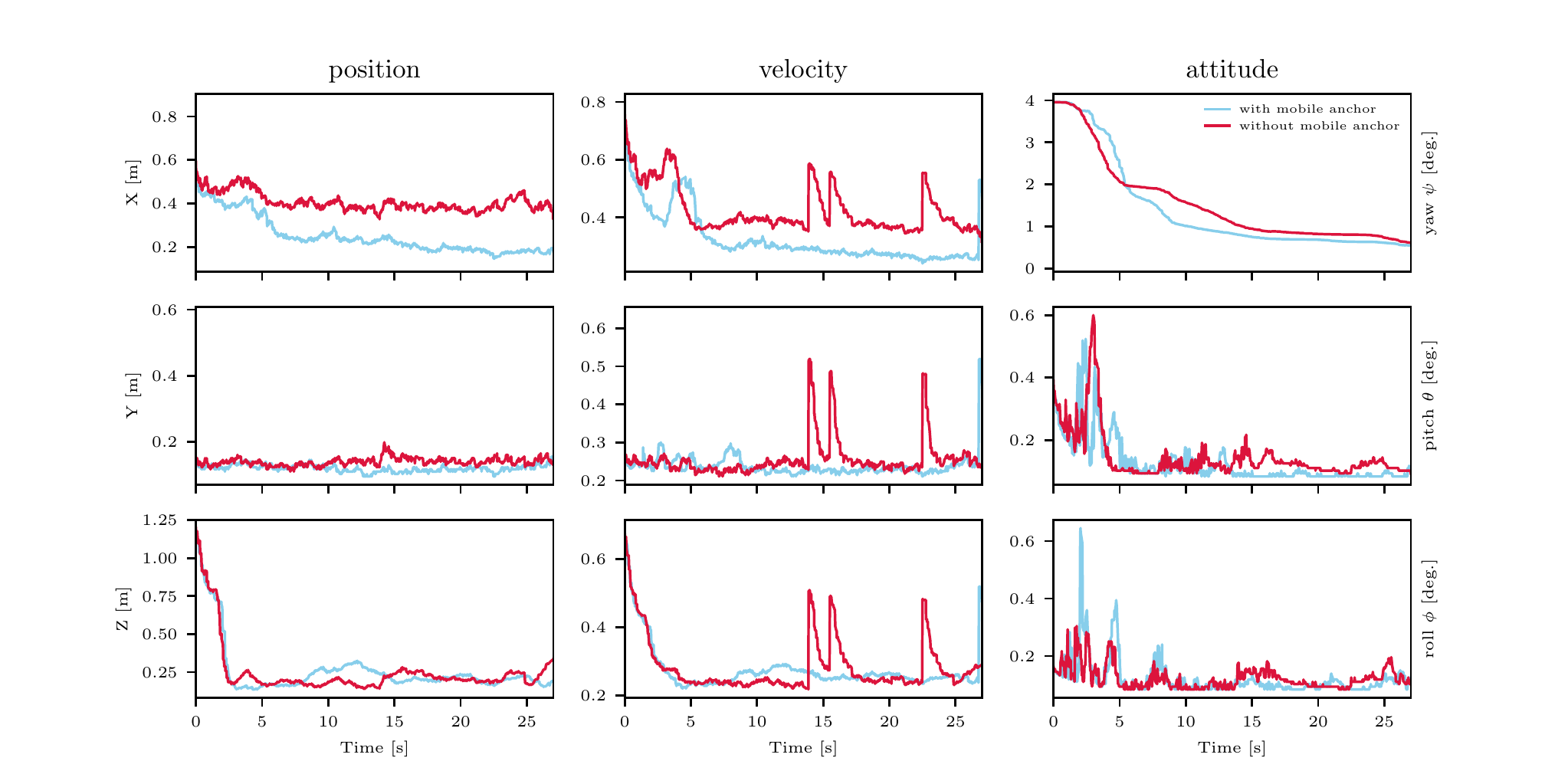}}
    \caption{Closed loop control of quadcopter using UWB ranging sensor with and without use of mobile anchor. The experiment illustrates the square roots of the diagonals of EKF covariance matrix (i.e. one standard deviation). As seen using mobile anchor, decreases uncertainty (position and velocity) dramatically in the $x$ direction, since there are enough fixed anchors along $y$ direction but nothing in the $x$ direction (see \figref{fig:anchors}). Note that the estimator does not compute the estimate in terms of yaw, pitch, and roll angles; the estimator output is simply transformed into this format as it is easier to parse in a figure.
    }
    \label{fig:compare_cov}
\end{figure*}
\begin{figure}
    \centering
    \makebox[0pt]{\includegraphics{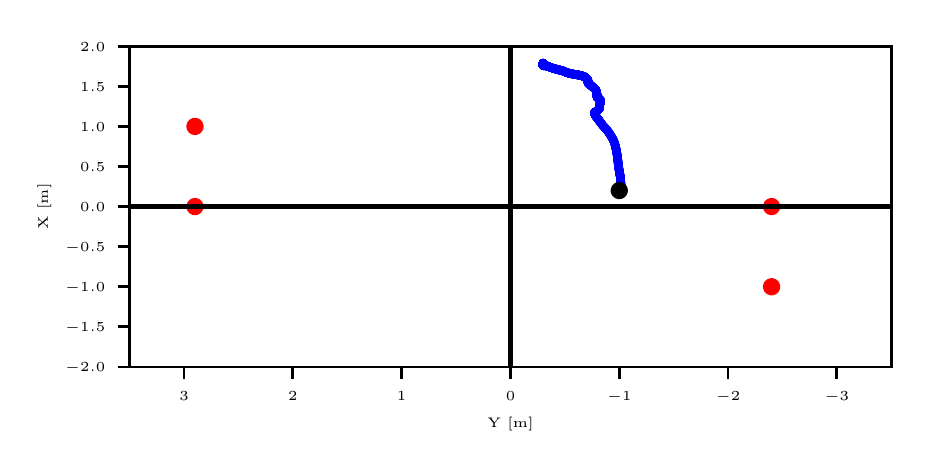}}
    \caption{Top view. Red dots represent the position of fixed anchors in both sets of experiments. The black dot is the fifth fixed anchor in the first and the initial position of mobile anchor in the second set of experiments. The set of four anchors are all located at $z=0$. The blue path is the  projection of the mobile anchor trajectory on $xy$ plane.     
    } 
    \label{fig:anchors}
\end{figure}
\subsection{Experimental Setup}
\begin{figure}
    \centering
    \includegraphics[width=0.5\linewidth]{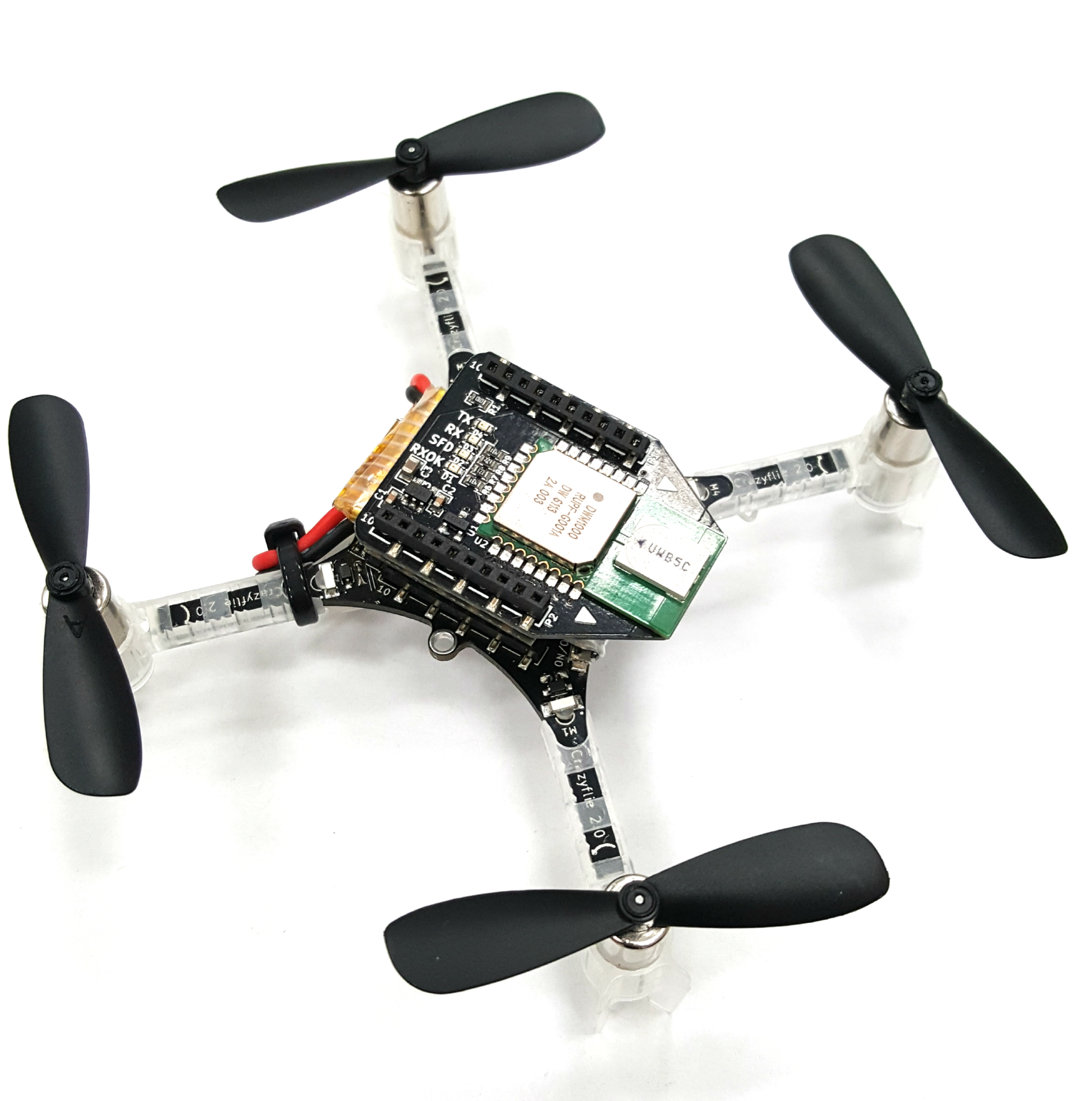}
    \caption{
    The quadcopter used in our experiments as the agent and mobile anchor.}
    \label{fig:cf}
\end{figure}

The testbed we used to examine our algorithm is a Crazyflie 2.0 quadcopter (shown in \figref{fig:cf}), with approximate mass of $30 \si{g}$, and motor-to-motor distance of $105 \si{mm}$. We used two quadcopters as the agent and the mobile anchor in our experiments. The quadcopter is equipped with an STM32F4 microcontroller, an Invensense MPU9250 inertial measurement unit, and a Decawave DW1000 radio module for the ultra-wideband ranging measurements. Beside the mobile anchor which shared the same hardware, all the fixed anchors also have the same computational and sensing hardware. The state estimation (EKF) for the agent was performed on the microcontroller, but the mobile anchor used motion capture system for its own localization, and was commanded by the trajectory according to \eqref{eq:strategy} computed off-board on a computer.    
Measurements from the accelerometer and rate gyroscope were taken at $500\si{Hz}$, and range measurements were taken at approximately $80\si{Hz}$. 
The effectiveness of our approach on the estimator performance is quantified by using the motion capture system, whose measurements are taken as ground truth. 
The top view of the anchor arrangement can be seen in \figref{fig:anchors}. Notable is that the anchors are placed such that their measurements 
carry sufficient information in the $y$ direction, but not very much in the $x$ direction. This was chosen so as to highlight the importance of mobile anchor and our proposed algorithm.

\subsection{Experiment with mobile anchor}
Two sets of similar experiments were conducted to verify the advantage of using mobile anchor against only fixed anchors to improve the localization accuracy. In the first set of experiments, we used five fixed anchors placed on the ground as pictured in \figref{fig:anchors}. We ran this test several times while the quadcopter hovers at the same position at origin and the EKF described in \secref{secEstimator} runs on the quacopter's microcontroller. In the second set of experiments, we used four fixed anchors in addition to one mobile anchor. The mobile anchor starts from its initial position (black dot in \figref{fig:anchors}) and moves according to the algorithm described in \secref{secMobileAlgorithm}.

\begin{figure}
    \centering
    \makebox[0pt]{\includegraphics{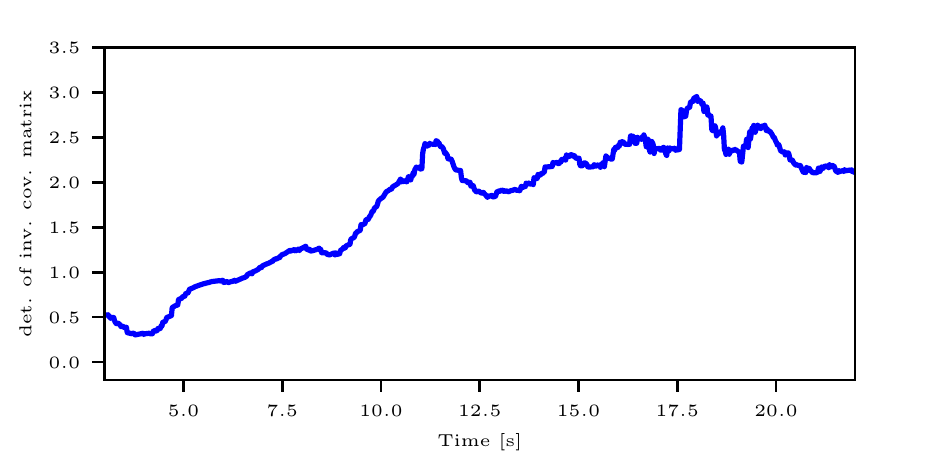}}
    \caption{Determinant of the inverse of the covariance matrix. The plot shows the increase of the determinant as expected when the mobile anchor moving in the direction of gradient ascent; that means the state estimate variance is decreasing.}
    \label{fig:det}
\end{figure}

Table \ref{tab:RMSE} compares the average root mean square error (RMSE) on state estimates of the quadcopter (position, velocity, attitude) from several trials for both sets of experiments. As seen the average RMSE is improved by about $14\%$, $9\%$, and $3\%$ for position, velocity and attitude respectively. The results show larger improvement on position rather than other states. This represent the fact that the proposed algorithm have direct impact on the position estimate quality but not on velocity and attitude.  
\figref{fig:det} shows the determinant of inverse covariance matrix in the proposed algorithm $\det(\mvec{A}^T\mvec{A})^{-1}$. This verifies that the mobile anchor has moved to decrease the position uncertainty around the agent.

\figref{fig:compare_cov} shows the experimental data of a trial of both experiments, which is the comparison of one standard deviation of state estimates. Comparing the monotonic decrease in the uncertainty of position in $x$ direction (blue line) with no improvement in $y$ shows that our intuition about anchor's arrangement was correct; absence of any anchor along the axis of symmetry between the fixed anchors causes insufficient available information along that axis, but when the mobile anchor starts to move towards the axis the uncertainty also decreases accordingly.
\begin{table}
\caption{Comparison of estimation error from experiments}
\begin{tabular}{ |p{1.6cm}|p{2cm}|p{2cm}|p{1.5cm}|  }
 \hline
 \multicolumn{4}{|c|}{RMSE} \\
 \hline
 &\multicolumn{2}{c|}{Average} & \multirow{2}{*}{difference (\%)} \\
 &4 fixed, 1 mobile&5 fixed anchors &\\
 \hline
 \hline
 position [m] & 0.156 & 0.182 & -14.3\%\\
 \hline
 velocity [m/s] & 0.374 & 0.409 & -8.6\%\\
 \hline
 attitude [deg.] & 5.140 & 5.315 & -3.3\%\\
 \hline
\end{tabular}
    \label{tab:RMSE}
\end{table}

\begin{figure}
    \centering
    \makebox[0pt]{\includegraphics{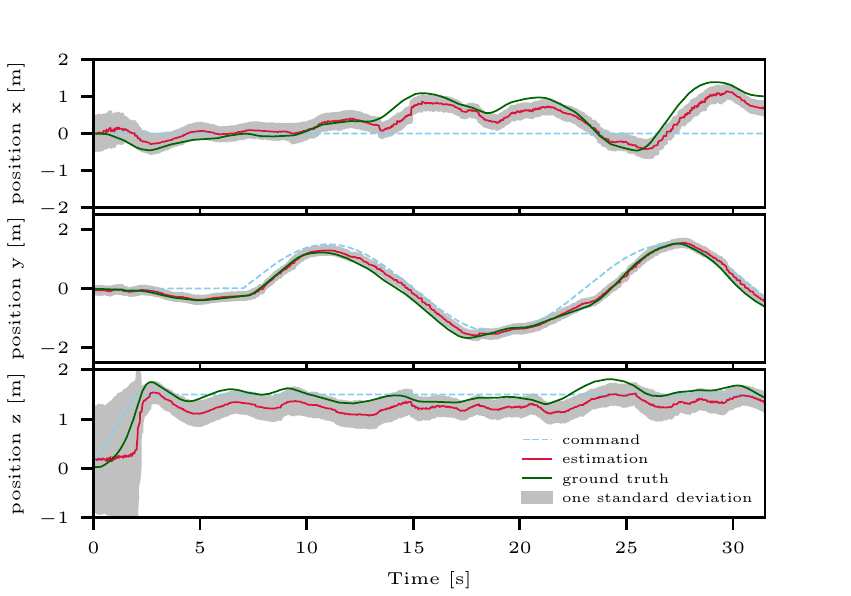}}
    \caption{Closed loop control of the quadcopter using one mobile anchor. The plot shows that the quadcopter is tracking a horizontal trajectory (dashed blue) along the $y$ axis. Despite the fact that the method was developed for a fixed agent, the results show that it can be used for moving agents as well.}
    \label{fig:trajectory}
\end{figure}

As the agent navigates in the environment to reach the target position, the localization network (anchors) is capable to reconstruct itself in real time accordingly. Although the proposed algorithm for the mobile anchor was derived assuming that the agent's position is fixed (i.e. no dynamic model was introduced), the algorithm works reliably even for the moving agent. As an example, \figref{fig:trajectory} shows an experiment when the quadcopter tracks a trajectory, and the mobile anchor adapts its position based on the proposed approach reducing the uncertainty of the quadcopter's position estimate.

\section{Conclusion}
\label{secConclusion}

The paper presented a EKF estimator for estimating a 6DOF states of an object using accelerometer, rate gyroscope, and UWB ranging sensor. Specifically, the paper was focused on developing a method to use UWB mobile anchor and improving the localization quality of an agent. This was done by minimizing the determinant of the covariance matrix. The minimization will result a set of closed form equations representing a direction that the mobile anchor moves in that direction at each time instant.

By using an UWB mobile anchor and the presented method, the quadcopter was able to reliably fly (hovering and tracking a trajectory) while the fixed anchors was set in specific locations to cause a large uncertainty in one direction which makes the flight very hard. The experimental results presented in this paper have shown a improvement on the position estimate (about $14\%$), which had also an indirect impact on the improvement of the velocity and attitude estimates. In addition, the result of a trajectory tracking scenario showed that the proposed method can be applied for the moving agents, and will result to reducing the overall uncertainty of the state estimates.

\section{Acknowledgement}
\label{secAcknowledgement}

This material is based upon work supported by the National Science Foundation Graduate Research Fellowship under Grant No. DGE 1106400.

\section*{References}
{
\printbibliography[heading=none, resetnumbers=true]
}

\end{document}